\documentclass[final]{cvpr}

\usepackage{times}
\usepackage{epsfig}
\usepackage{graphicx}
\usepackage{amsmath}
\usepackage{amssymb}
\usepackage{symbols}
\usepackage{arydshln}
\usepackage[table,dvipsnames]{xcolor}
\usepackage{xspace}
\usepackage{capt-of,etoolbox}
\usepackage{caption}
\usepackage{subcaption}
\usepackage{booktabs}
\usepackage{enumitem}
\usepackage{pifont}

\usepackage[pagebackref=true,breaklinks=true,colorlinks,bookmarks=false]{hyperref}

\newcommand{\thor}{\textsc{AI2-THOR}\xspace}
\newcommand{\iou}{\text{IOU}\xspace}

\newcommand{\cmark}{\ding{51}}%
\newcommand{\xmark}{\ding{55}}%
\newcommand{\Onephase}{\emph{1-Phase}\xspace}%
\newcommand{\Twophase}{\emph{2-Phase}\xspace}%

\newcommand{\fixedstrict}{\textsc{\%FixedStrict}\xspace}%
\newcommand{\success}{\textsc{Success}\xspace}%
\newcommand{\energy}{\textsc{\%E}\xspace}%
\newcommand{\numchanged}{\textsc{\#Changed}\xspace}%

\def\cvprPaperID{5618} 
\def\confYear{CVPR 2021}

\begin{document}

\title{Visual Room Rearrangement}

\author{Luca Weihs${^1}$\\
\and Matt Deitke${^{1,2}}$ \and Aniruddha Kembhavi${^{1,2}}$ \and Roozbeh Mottaghi${^{1,2}}$
}

\twocolumn[{
\renewcommand\twocolumn[1][]{#1}
\maketitle
\vspace*{-0.15in}
\centering
\includegraphics[width=.9\linewidth]{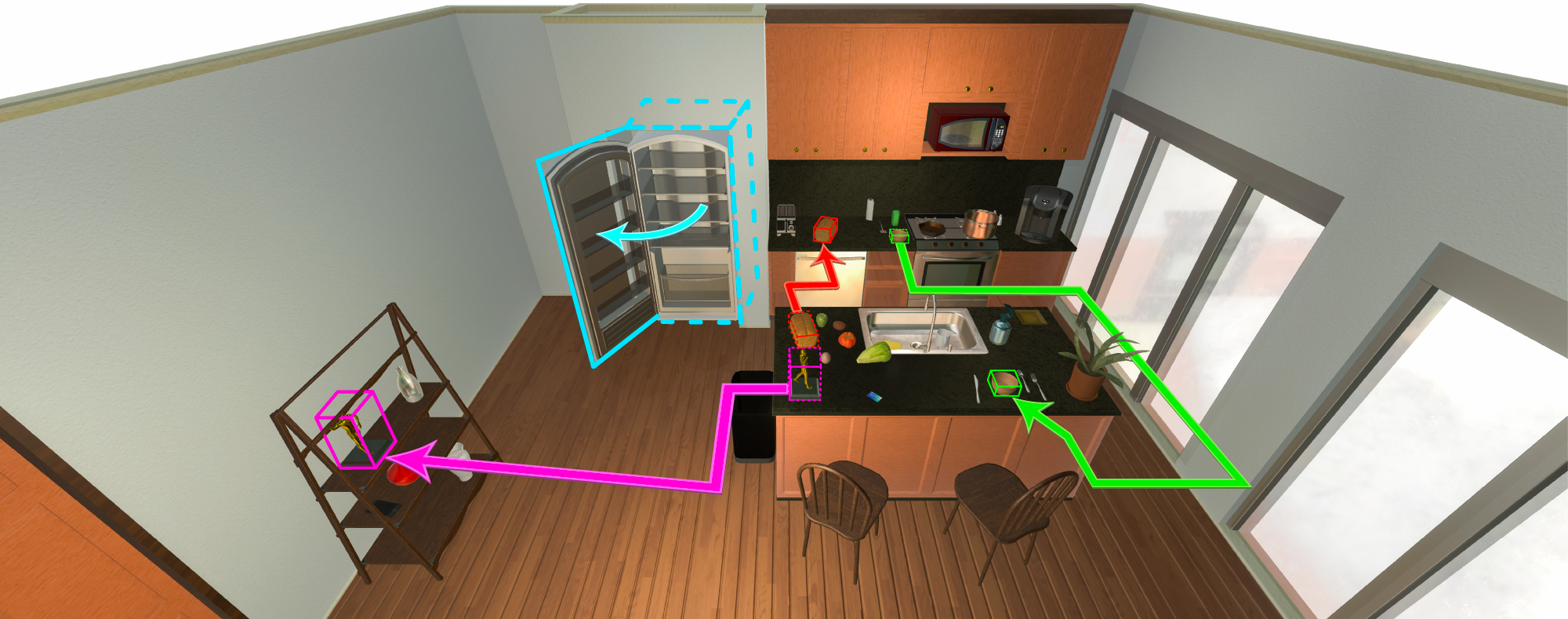}
\\[-0.05in]
\captionof{figure}{An instance of the Room Rearrangement task. Objects begin in the positions indicated by the solid 3D bounding boxes. An agent must walk through the room and record the objects it sees. The agent is then removed, and objects are moved to the locations indicated by the dashed bounding boxes. The agent is then reintroduced into the room and must interact with objects (moving or opening them) to return the room to its original state.\\}
\label{fig:teaser}
}]

\maketitle
\thispagestyle{empty}
\begin{abstract}
\vspace{-0.3cm}
There has been a significant recent progress in the field of Embodied AI with researchers developing models and algorithms enabling embodied agents to navigate and interact within completely unseen environments. In this paper, we propose a new dataset and baseline models for the task of Rearrangement. We particularly focus on the task of Room Rearrangement: an agent begins by exploring a room and recording objects' initial configurations. We then remove the agent and change the poses and states (e.g., open/closed) of some objects in the room. The agent must restore the initial configurations of all objects in the room. Our dataset, named RoomR, includes 6,000 distinct rearrangement settings involving 72 different object types in 120 scenes. Our experiments show that solving this challenging interactive task that involves navigation and object interaction is beyond the capabilities of the current state-of-the-art techniques for embodied tasks and we are still very far from achieving perfect performance on these types of tasks.
\end{abstract}


\section{Introduction}
One of the longstanding goals of Embodied AI is to build agents that interact with their surrounding world and perform tasks. Recently, navigation and instruction following tasks have gained popularity \cite{anderson18,Anderson2018VisionandLanguageNI,Batra2020ObjectNavRO} in the Embodied AI community. These tasks are the building blocks of interactive embodied agents, and over the past few years, we have observed remarkable progress regarding the development of models and algorithms. However, a typical assumption for these tasks is that the environment is static; namely, the agent can move within the environment but cannot interact with objects or modify their state. The ability to interact with and change its environment is crucial for any artificial embodied agent and cannot be studied in static environments. There is a general trend towards interactive tasks \cite{sapien,ALFRED20,xiali2020relmogen}. These tasks focus on specific aspects of interaction such as object manipulation, long-horizon planning and understanding pre-condition and post-conditions of actions. In this paper, we address a more comprehensive task in a visually rich environment that can subsume each of these skills. 

We address an instantiation of the \emph{rearrangement} problem, an interactive task, recently introduced by Batra et al. \cite{batra2020rearrangement}.
The goal of the rearrangement task is to reach a goal room configuration from an initial room configuration through interaction. In our instantiation, an agent must recover a scene configuration after we have randomly moved, or changed the state of, several objects (\eg see Fig.~\ref{fig:teaser}).

This problem has two stages: \emph{walkthrough} and \emph{unshuffle}. During the walkthrough stage, the agent may explore the scene and, through egocentric perception, record information regarding the goal configuration. We then remove the agent from the room and move some objects to other locations or change their state (\eg opening a closed microwave). In the unshuffle stage, the agent must interact with objects in the room to recover the goal configuration observed in the walkthrough stage.

Rearrangement poses several challenges such as inferring the visual differences between the initial and goal configurations, inferring the objects' state, learning the post-conditions and pre-conditions of actions, maintaining a persistent and compact memory representation during the walkthrough stage, and successful navigation.
To establish baseline performance for our task, we evaluate an actor-critic model akin to the state-of-the-art models used for long-horizon tasks such as navigation. 
We train our baselines using decentralized distributed proximal policy optimization (DD-PPO)~\cite{Wijmans2020DDPPOLN,schulman2017proximal}, a reward-based RL algorithm, as well as with DAgger~\cite{dagger}, a behavioral cloning method. During the walkthrough stage, the agent uses a non-parametric mapping module to memorize its observations along with any visible objects and their positions. In the unshuffle stage the agent compares images that it observes against what it has observed in its map and may use this information to inform which objects it should move or open. As a proof-of-concept we also run experiments with a model that includes a semantic mapping component adapted from the Active Neural SLAM model~\cite{chaplot2020learning}.

To facilitate research in this challenging direction, we compiled the Room Rearrangement (RoomR) dataset. RoomR is built upon AI2-THOR \cite{ai2thor}, a virtual interactive environment that enables interacting with objects and changing their state. The RoomR dataset includes 6{,}000 rearrangement tasks that involve changing the pose and state of multiple objects within an episode. The level of the difficulty of each episode varies depending on the differences between the initial and the goal object configurations. We have used 120 rooms and more than 70 unique object categories to create the dataset.

We consider two variations of the room rearrangement task. In the first setting, which we call the \Onephase task, the agent completes the walkthrough and unshuffle stages in parallel so that it is given aligned images from the walkthrough and unshuffle configurations at every step. In the second setting, the \Twophase task, the agent must complete the walkthrough and unshuffle stages sequentially; this \Twophase variant is more challenging as it requires the agent to reason over longer time spans. Highlighting the difficulty of the rearrangement, our evaluations show that our strong baselines struggle even in the easier \Onephase task. 
Rearrangement poses a new set of challenges for the embodied-AI community.
Our code and dataset are publicly available. A supplementary video\footnote{\url{https://youtu.be/1APxaOC9U-A}} provides the description of the task and some qualitative results.

\section{Related Work}
\noindent \textbf{Embodied AI tasks.} In recent years, we have witnessed a surge of interest in learning-based Embodied AI tasks. Various tasks have been proposed in this domain: navigation towards objects \cite{Batra2020ObjectNavRO,Yang2018VisualSN,Wortsman2019LearningTL,chaplot2020object} or towards a specific point \cite{anderson18,Savva_2019_ICCV,Wijmans2020DDPPOLN,ramakrishnan2020occant}, scene exploration \cite{chen2018learning,chaplot2020learning}, embodied question answering \cite{gordon18,embodiedqa}, task completion \cite{Zhu2017VisualSP}, instruction following \cite{Anderson2018VisionandLanguageNI,ALFRED20}, object manipulation \cite{corl2018surreal,pmlr-v100-yu20a}, multi-agent coordination \cite{twobodyproblem,cordialsync}, and many others. Rearrangement can be considered as a broader task that encompasses skills learned through these tasks. 

\noindent \textbf{Rearrangement.} \emph{Rearrangement Planning} is an established field in robotics research where the goal is to reach a goal state from an initial state \cite{BenShahar1996PracticalPP,stilman2007manipulation,king2016rearrangement,kron16,yuan2018rearrangement,labbe2020monte}. While these methods have shown impressive performance, they consider complete observability of the state from perfect visual perception \cite{cosgun2011push,king2016rearrangement}, a planar surface as the environment \cite{krontiris2015dealing,song2019multi}, a static robot \cite{dogar2011framework,krontiris2014rearranging}, same environment for evaluation of generalization \cite{scholz2010combining,King-2015-5955}, or a limited set of object categories or limited variation within the categories \cite{amazon,gualtieri2018pick}. Some works address some of these issues, such as generalization to new objects or imperfect perception \cite{zakka2020form2fit,berscheid2020self}. In this paper, we take a step further and relax these assumptions by considering raw visual input instead of perfect perception, a visually and geometrically complex scene as the configuration space, separate scenes for training and evaluation, a variety of objects, and object state changes. 

\noindent \textbf{Task and motion planning.} Our work can be considered as an instance of joint task and motion planning \cite{kaelbling2011hierarchical,srivastava2014combined,plaku2010sampling,garrett2018ffrob,dantam2016incremental} since solving the rearrangement task requires low-level motion planning to plan a sequence of actions and high-level task planning to recover the goal state from the initial state of the scene. However, the focus of these works is primarily on the planning problem rather than perception.

\section{The Room Rearrangement Task} \label{sec:task-def}

\subsection{Definition}

Our goal is to rearrange an initial configuration of a room into a goal configuration. So that our agent does not have to reason about soft-body physics, we restrict our attention to piece-wise rigid objects. Suppose a room contains $n$ piece-wise rigid objects. We define the state for object $i$ as $s_i = (p_i, o_i, c_i, b_i)$ where
\begin{enumerate}[label=$\bullet$,topsep=0pt,itemsep=-1ex,partopsep=1ex,parsep=1ex]
    \item $p_i \in \{\text{3D rotations and translations}\}=\text{SE}(3)$ records the pose of the object, 
    \item if the object can be opened $o_i\in[0,1]$ specifies the \emph{openness} of an object (\eg $o_i=0.5$ means a door is half open) and if the object cannot be opened (\eg a mug) then $o_i=\emptyset$,
    \item $c_i$ records the $8$ coordinates in $\bR^3$ of the corners of the 3D bounding box for object $i$, and
    \item $b_i\in\{0,1\}$ records if the $i$th object is ``broken'' (1 if broken, otherwise 0). 
\end{enumerate}
While this definition of an object's state is constrained (\eg objects can be more than just ``broken'' and ``unbroken'') it matches well the capabilities of our target embodied environment (AI2-THOR) and can be easily enriched as embodied environments become increasingly realistic. We now let $S=\text{SE}(3)\times ([0,1]\cup \{\emptyset\})\times \bR^{8\cdot 3}\times\{0,1\}$ be the set of all possible poses for a single object and $\cS=\prod_{i=1}^nS$
the set of all possible joint object poses. The agent's goal is to convert an initial configuration $s^0 \in \mathcal{S}$ to a goal $s^* \in \mathcal{S}$.

Our task has two stages: (1) \emph{walkthrough} and (2) \emph{unshuffle}. During the walkthrough stage, the agent is placed into a room with goal state $s^*$, and it should collect as much information as needed for that particular state of the room in a maximum number of actions (for us, 250).
The agent is removed from the room after the walkthrough stage. We then select a random subset of the $n$ objects and change their state. The state change may be a change in $p$ or $o$. This state will be the initial state $s^0$ that the agent observes at the beginning of the unshuffle stage. The agent's goal is to convert $s^0$ to $s^*$ ($s^0\rightarrow s^*$) via a sequence of actions.

\subsection{Metrics}\label{sec:metrics}

To quantify an agent's performance, we introduce four metrics below. Recall from the above that an agent begins an unshuffle episode with the room in state $s^0$ and has the goal of rearranging the room to end in state $s^*$. Suppose that at the end of an unshuffle episode, the agent has reconfigured the room so that it lies in state $s=(s_1,\ldots,s_n)\in\cS$. In practice, we cannot expect that the agent will place objects in exactly the same positions as in $s^*$. We instead choose a collection of thresholds which determine if two object poses are, approximately, equal. When two poses $(s_i,\,s^*_i)$ are approximately equal we write $s_i\approx s^*_i$. Otherwise we write $s_i\not\approx s^*_i$.

Let $s_i^1,s_i^2\in S$ be two possible poses for object $i$. As it makes little intuitive sense to compare the poses of broken objects, we will always assert that poses of broken objects are unequal. Thus if $b_i^1 = 1$ or $b_i^2=1$ we define $s^1_i\not\approx s^2$. Now let's assume that neither $b_i^1 = 1$ nor $b_i^2=1$. If object $i$ is pickupable, let $\iou(s_i^1,s_i^2)$ be the intersection over union between the 3D bounding boxes $c^1_i,c^2_i$. We then say that $s^1_i\approx s^2_i$ if, and only if, $\iou(s_i^1,s_i^2)\geq 0.5$. If object $i$ is openable but not pickupable, we say that $s^1_i\approx s^2$ if, and only if, $|o_i^1-o_i^2| \leq 0.2$. The use of the IOU above means that object poses can be approximately equal even when their orientations are completely different. While this can be easily made more stringent, our rearrangement task is already quite challenging. Note also that our below metrics do not consider the case where there are multiple identical objects in a scene (as this does not occur in our dataset). We now describe our metrics.

\noindent\textbf{Success} (\success) -- This is the most unforgiving of our metrics and equals 1 if all object poses in $s$ and $s^*$ are approximately equal, otherwise it equals 0.

\noindent\textbf{\% Fixed (Strict)} (\fixedstrict) -- The above \success metric does not give any credit to an agent if it manages to rearrange some, but not all, objects within a room. To this end, let $M_{\text{start}}=\{i\mid s^0_i \not\approx s^*_i\}$ be the set of misplaced objects at the start of the unshuffle stage and let $M_{\text{end}}=\{i\mid s_i \not\approx s^*_i\}$ be the set of misplaced objects at the end of the episode. We then let \fixedstrict equal 0 if $|M_{\text{end}} \setminus M_{\text{start}}| > 0$ (\ie the agent has moved an object that should not have been moved)
and, otherwise, let \fixedstrict equal $1 - |M_{\text{end}}| / |M_{\text{start}}|$ (\ie the proportion of objects that were misplaced initially but ended in the correct pose).

\noindent\textbf{\% Energy Remaining} (\energy) -- Missing from all of the above metrics is the ability to give partial credit if, for example, the agent moves an object across a room and towards the goal pose, but fails to place it so that it has a sufficiently high IOU with the goal. To allow for partial credit, we define an energy function $D:S\times S\to[0,1]$ that monotonically decreases to 0 as two poses get closer together (see the Appendix~\ref{app:energy} for full details) and which equals zero if two poses are approximately equal. The \energy metric is then defined as the amount of energy remaining at the end of the unshuffle episode divided by the total energy at the start of the unshuffle episode, \eg $\textsc{\%E} = (\sum_{i=1}^n D(s_i, s^*_i)) / (\sum_{i=1}^n D(s^0_i, s^*_i))$.

\noindent\textbf{\# Changed} (\numchanged) -- To give additional insight as to our agent's behavior we also include the \numchanged metric. This metric is simply the the number of objects whose pose has been changed by the agent during the unshuffle stage. Note that larger or smaller values of this metric are not necessarily ``better'' (both moving no objects and moving many objects randomly are poor strategies).

The above metrics are then averaged across episodes when reporting results.

\section{The RoomR Dataset}\label{sec:dataset}

The Room Rearrangement (RoomR) dataset utilizes 120 rooms in AI2-THOR \cite{ai2thor} and contains 6,000 unique rearrangements (50 rearrangements per training, validation, and testing room). Each datapoint consists of an initial room state $s_0$, the agent's starting position, and the goal state $s^*$.

\subsection{Generating Rearrangements}

\noindent
The automatic generation of the dataset enables us to scale up the number of rearrangements easily. We generate each room rearrangement using the procedure that follows.

\noindent \textbf{Place agent.} We randomize the agent's position on the floor. The position is restricted to lie on a grid, where each cell is of size $0.25\mathrm{m}\times 0.25\mathrm{m}$. The agent's rotation is then randomly chosen amongst $\{0^\circ, 90^\circ, 180^\circ, 270^\circ\}$. The agent's starting pose is the same for both $s^0$ and $s^*$.

\noindent \textbf{Shuffle background objects.} To obtain different configurations of objects for each task in the dataset, we randomly shuffle each movable object, ensuring background objects do not always appear in the same position. Shuffled objects are never hidden inside other receptacles (\eg fridges, cabinets), which reduces the task's complexity.

\noindent \textbf{Sample objects.} We now randomly sample a set of $N\geq 0$ openable but non-pickupable objects and a set of $M\geq 0$ pickupable objects. These objects and counts are chosen randomly with $N\in\{0,1\}$ and $M\in\{1-N, ..., 5-N\}$.

\noindent \textbf{Goal ($s^*$) setup.} We open the $N$ objects sampled in the last step to some randomly chosen degree of openness in $[0,1]$ and move the other $M$ pickupable objects to arbitrary locations within the room. The room's current state is now $s^*$, the start state for the walkthrough stage.

\noindent\textbf{Initial ($s^0$) setup.} We randomize the $N$ sampled openable objects' openness and shuffle the position of each of the $M$ sampled pickupable objects once more. We are now in $s^0$, the start state for the unshuffle stage.

\noindent
In the above process, we ensure that no broken objects are in $s^0$ or $s^*$. While we provide a fixed number of datapoints per room, this process can be used to sample a practically unbounded number of rearrangements.

\subsection{Dataset Properties}

\begin{figure}[t!]
    \centering
    \includegraphics[width=0.47\textwidth]{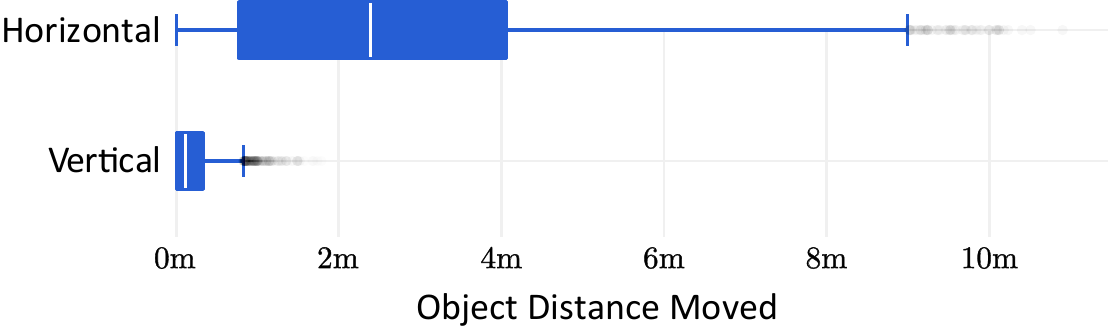}
    \caption{\textbf{Distance distribution.} The horizontal (Manhattan distance) and vertical distance distributions between changed objects in their goal and initial positions.}
    \vspace{-0.4cm}
    \label{fig:distance}
\end{figure}

\noindent\textbf{Rooms.}
There are 120 rooms across the categories of kitchen, living room, bathroom, and bedroom (30 rooms for each category). We designate 20 rooms for training, 5 rooms for validation, and 5 rooms for testing, across each room category. Of the 6${,}$000 unique rearrangements in our dataset, 4000 are designated for training, 1000 are set in validation rooms, and 1000 are set in test rooms. For each such split, there are 50 rearrangements per room.

\noindent\textbf{Objects.}
There are 118 object categories (listed in Appendix~\ref{app:objects}), among which 62 are pickupable (\eg cup) and 10 are openable and non-pickupable (\eg fridge). The set of object categories that appear in the validation and testing rooms is a subset of the object categories that appear during training. Thus, if a plant appears in a validation or testing room, then a plant is also present in one of the training rooms. While all object categories are seen during training, the physical appearance of object instances are often unique in training, validation, and testing rooms. \thor provides annotation as to if an object is pickupable, openable, movable, or static.

\begin{figure}[t!]
    \centering
    \includegraphics[width=0.47\textwidth]{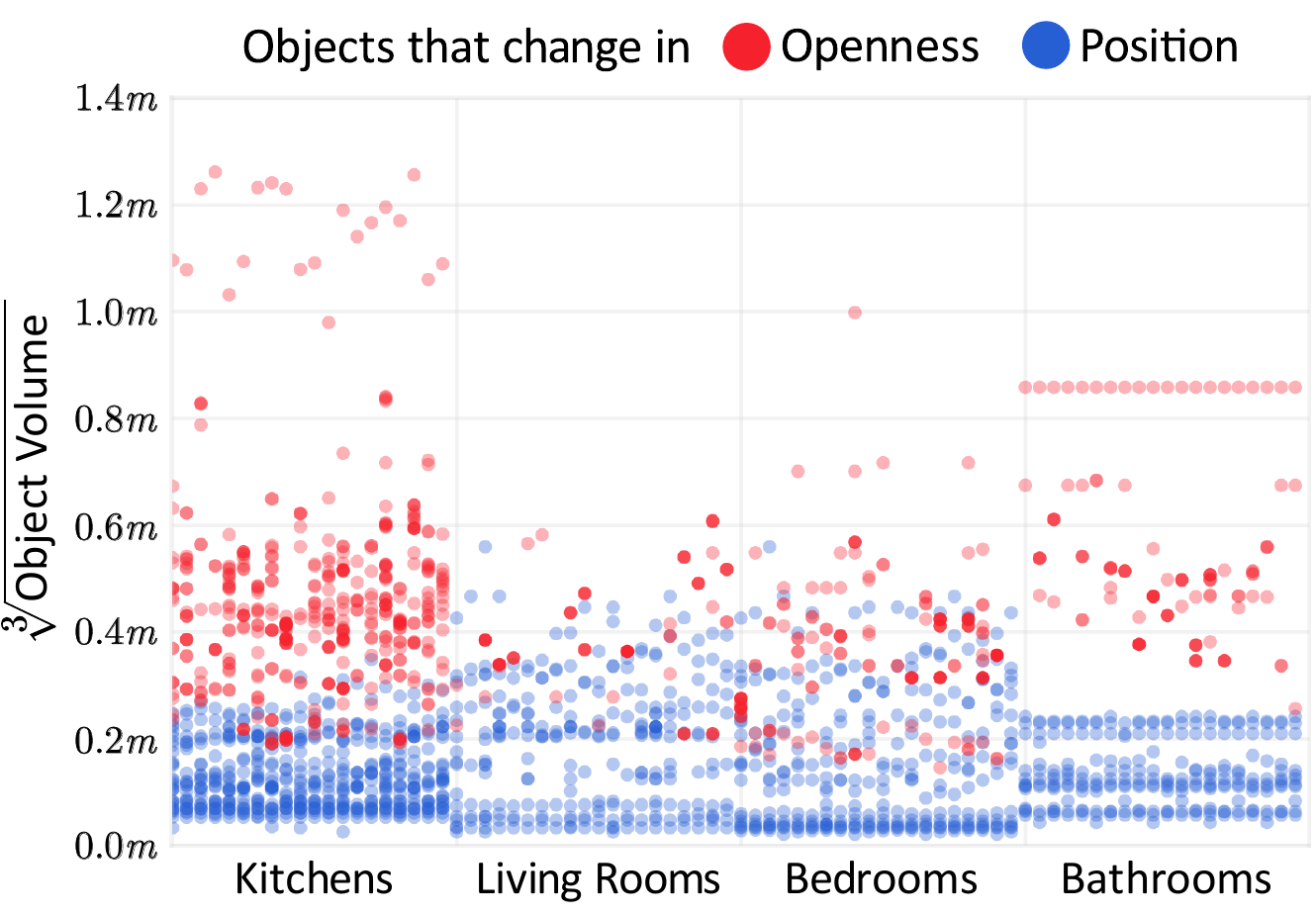}
    \caption{\textbf{Distribution of object size.} Each column contains the cube root of every object's bounding box volume that may change in openness (red) or position (blue) for a particular room. Notice that, across room categories, objects that change in position are significantly smaller than objects that change in openness.}
        \vspace{-0.4cm}
    \label{fig:volumes}
\end{figure}

Across the dataset, there are 1895 pickupable object instances and 1262 openable non-pickupable object instances (an average of 15.7 and 10.5, respectively, per room). Fig.~\ref{fig:distance} shows the distance distribution (horizontal and vertical) of objects between their initial and goal positions. It illustrates the complexity of the problem, where the agent must travel relatively far to recover the goal configuration. Fig.~\ref{fig:volumes} shows the distribution of these object groups and their sizes within every room. Note that pickupable objects (\eg apple, fork) tend to be relatively small and hard to find, compared to openable non-pickupable objects (\eg cabinets, drawers). Further, across room categories, the number of openable non-pickupable objects varies considerably.

\section{Model}\label{sec:model}

\begin{figure*}[tp]
    \centering
    \includegraphics[width=0.8\textwidth]{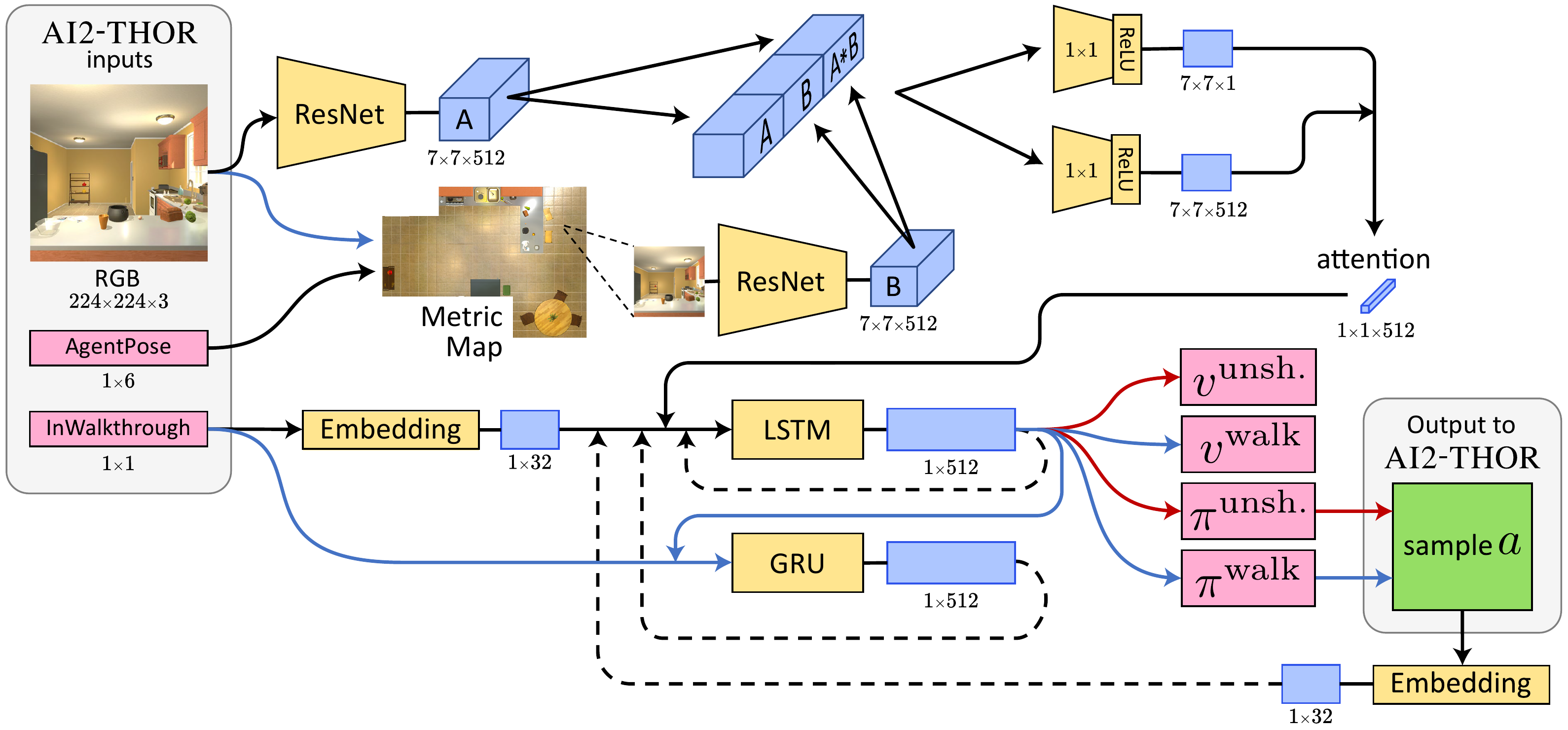}
    \vspace{-0.3cm}
    \caption{\textbf{Model overview.} The model is used for both the unshuffle and walkthrough stages. The connections specific to the walkthrough and unshuffle stages are shown in blue and red, respectively. The dashed lines represent connections from the previous time step. The model's trainable parameters, inputs and outputs, and intermediate features are shown in yellow, pink, and blue, respectively.}
    \vspace{-0.3cm}
    \label{fig:model}
\end{figure*}

In our experiments, Sec.~\ref{sec:experiments}, we consider two RoomR task variants: \Onephase and \Twophase. In the \Onephase task, the agent completes the unshuffle and walkthrough stages simultaneously in lock step. The model we employ for this \Onephase task is a simplification of the model used when performing the \Twophase task (in which both stages must be completed sequentially and so longer-term memory is required). For space we only describe the \Twophase model below, see our codebase for all architectural details.

Our network architecture, see Fig.~\ref{fig:model}, follows the same basic structure as is commonly employed within Embodied AI tasks \cite{Savva_2019_ICCV,Wijmans2020DDPPOLN,robothor,WeihsKembhaviEtAl2019,cordialsync}: a combination of a convolutional neural network to process input egocentric images, a collection of embedding layers to encode discrete inputs, and an RNN to enable the agent to reason through time. In addition to this baseline architecture, we would like our agent to have two capabilities relevant to the rearrangement task, namely the abilities to, during the unshuffle stage, (a) explicitly compare images seen during the walkthrough stage against those seen during the unshuffle stage, and (b) reference an implicit representation of the walkthrough stage. We now describe the details of our architecture and how we enable these additional capabilities.

Our agents are of the actor-critic~\cite{Mnih2016AsynchronousMF} variety and thus, at each timestep $t\geq 0$, given observations $\omega_t$ (\eg an egocentric RGB image) and a summary $h_{t-1}$ of the agent's history, we require that an agent produces a policy $\pi_\theta(\omega_t\mid h_{t-1})$ (\ie a distribution over the agent's actions) and a value $v_\theta(\omega_t\mid h_{t-1})$ (\ie an estimate of future rewards). Here we let $\theta\in\Theta$ be a catch-all parameter representing all of the trainable parameters in our network. As we wish for our agent to have characteristically different behavior in the walkthrough and unshuffle stages, we have two separate policies $\pi^{\text{walk}}_\theta$ and $\pi^{\text{unsh.}}_\theta$ (and similarly for $v_\theta$).

To encode input 224${\times}$224${\times}$3 RGB egocentric images, we use a ResNet18~\cite{resnet} model (pretrained on ImageNet) with frozen model weights with the final average pooling and classification layers removed. This ResNet18 model transforms input images into $7{\times}7{\times}512$ tensors. For our RNN, we leverage a 1-layer LSTM~\cite{HochreiterNC1997} with 512 hidden units. To produce the policies $\pi^{\text{walk}}$ and $\pi^{\text{unsh.}}$ we use two $512\times 84$ linear layers, each applied to the output from the LSTM, and each followed by a softmax nonlinearity. Similarly, to produce the two values $v^{\text{walk}}$ and $v^{\text{unsh.}}$ we use two distinct $512\times 1$ linear layers applied to the output of the LSTM with no additional nonlinearity. We now describe how we enable agents the abilities (a) and (b) above.

\noindent\textbf{Mapping and image comparison.} Our model includes a non-parametric mapping module. The module saves the RGB images seen by the agent during the walkthrough stage, along with the agent's pose. During the unshuffle stage, the agent (i) queries the metric map for all poses visited during the walkthrough stage, (ii) chooses the pose closest to the agent's current pose, and then (iii) retrieves the image saved by the walkthrough agent at that pose. Using an attention mechanism, the agent can then compare this retrieved image against its current observation to decide which objects to target.

\noindent\textbf{Implicit representations of the walkthrough stage.} In addition to explicitly storing the images seen during the walkthrough stage, we also wish to enable our agent to produce an implicit representation of its experiences during the walkthrough stage. To this end, at every timestep $t$ during the walkthrough stage we pass $h_t$, the output of the 1-layer LSTM described above, to a 1-layer GRU with 512 hidden units to produce the walkthrough encoding $w_t$. During the unshuffle stage this walkthrough encoding is no longer updated and is simply taken as the encoding from the last walkthrough step. The walkthrough encoding is passed as an input to the LSTM in a recurrent fashion.

\section{Experiments}\label{sec:experiments}
\vspace{-0.2cm}
This section provides the results for several baseline approaches that achieve state-of-the-art performance on other embodied tasks (\eg navigation). The room rearrangement task and the RoomR dataset are very challenging. To make the problem more manageable, we simplify assumptions in choosing the action space and the sensor modalities. Sec. \ref{sec:roomr-actions} and Sec. \ref{sec:roomr-variants} explain the details of the action space and sensor modalities, respectively. We show that even with these simplifications, the baseline models struggle.

\subsection{Action Space}\label{sec:roomr-actions}
\vspace{-0.2cm}
AI2-THOR offers a wide variety of means by which agents may interact with their environment ranging from ``low-level'' (\eg applying forces to individual objects) to ``high-level'' (\eg open an object of the given type) interactions. Prior work, \eg \cite{twobodyproblem,WeihsKembhaviEtAl2019,cordialsync,gordon18,ALFRED20} has primarily used higher-level actions to abstract away some details that would otherwise distract from the problem of interest. We follow this prior work and define our agent's action space as $\cA = \cA_{\text{Nav.}} \cup \cA_{\text{Rotate}} \cup \cA_{\text{Look}} \cup \cA_{\text{UpDown}} \cup \cA_{\text{Pickup}} \cup \cA_{\text{Open}} \cup \{\textsc{PlaceObject}, \textsc{Done}\}$ where taking action:
\begin{itemize}[leftmargin=*]
    \itemsep0em 
    \item $a\in \cA_{\text{Nav.}} = \{\textsc{MoveX} \mid \textsc{X} \in \{$\textsc{Ahead}, \textsc{Left}, \textsc{Right}, \textsc{Back}$\}\}$ results in the agent moving 0.25m in the direction specified by $\textsc{X}$ in the agent's coordinate frame (unless this would result in the agent colliding with an object).
    \item $a\in \cA_{\text{Rotate}} = \{\textsc{RotateLeft}, \textsc{RotateRight}\}$ results in the agent rotating 90$^\circ$ clockwise if $a=\textsc{RotateRight}$ and 90$^\circ$ counter-clockwise if $a=\textsc{RotateLeft}$.
    \item $a\in \cA_{\text{Look}} = \{\textsc{LookUp}, \textsc{LookDown}\}$ results in the agent lowering/raising its camera angle by 30$^\circ$,
    \item $a\in \cA_{\text{Pickup}} = \{\textsc{PickupX} \mid \textsc{X} \in \{$the 62 pickupable object types$\}\}$ results in the agent picking up a visible object of type \textsc{X} if: (a) the agent is not already holding an object, (b) the agent is close enough to the object (within 1.5m), and (c) picking up the object would not result in it colliding with objects in front of the agent. If there are multiple objects of type \textsc{X} then the closest is chosen.
    \item $a\in \cA_{\text{UpDown}}=\{\textsc{Stand}, \textsc{Crouch}\}$ results in the agent raising or lowering the agent's camera to one of two fixed heights allowing it to, \eg, see objects under tables.
    \item $a\in \cA_{\text{Open}} = \{\textsc{OpenX} \mid \textsc{X} \in \{$the 10 openable object types that are not pickupable$\}\}$, if an object whose openness is different from the openness in the goal state is visible and within 1.5m of the agent, this object's openness is changed to its value in the goal state.
    \item $a=\textsc{PlaceObject}$ results in the agent dropping its held object. If the held object's goal state is visible and within 1.5m of the agent, it is placed into that goal state. Otherwise, a heuristic is used to place the object on a nearby surface.
    \item $a=\textsc{Done}$ results in the walkthrough or unshuffle stage immediately terminating.
\end{itemize}
In total, there are $|\cA|=84$ possible actions. Some of the above actions have been designed to be fairly abstract or ``high-level,'' \eg the \textsc{PlaceObject} action abstracts away all object manipulation complexities. As we discuss in Appendix~\ref{app:lowlevel}, we have implemented ``lower-level'' actions. Still, we stress that, even with these more abstract actions, the planning and visual reasoning required in RoomR already makes the task very challenging.

\subsection{RoomR Variants}\label{sec:roomr-variants}
\vspace{-0.2cm}
\begin{figure*}[t!]
    \vspace{-0.3cm}
    \centering
    \includegraphics[width=0.95\textwidth]{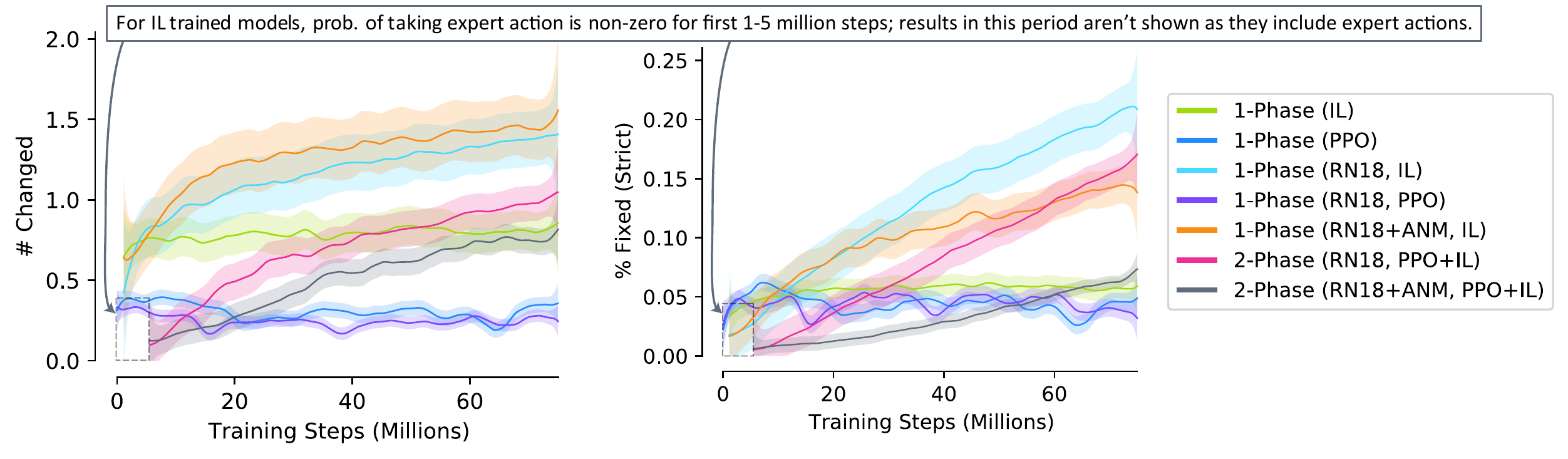}
    \vspace{-0.5cm }
    \caption{\textbf{Performance over training.} The (training-set) performance of our models over  ${\sim}75$Mn training steps. We report the \numchanged and \fixedstrict metrics, shown values and 95\% error bars are generated using locally weighted scatterplot smoothing. Notice that the PPO models quickly saturate suggesting that they become stuck in local optima. IL continue to improve throughout training although Tab.~\ref{tab:results} suggests that these models begin to overfit on the training scenes.}
    \label{fig:training-curves}
\end{figure*}

We will now detail the \Onephase and, more difficult, \Twophase variants of our RoomR task. These variants are, in part, defined by the sensors available to the agent. We begin by listing all sensors (note that only a subset of these will be available to any given agent in the below variants): 

\begin{table*}[tp]
\centering

\footnotesize
\vspace{-0.2cm}
\begin{tabular}{l|ccc|ccc|ccc|ccc}
\toprule
{} & \multicolumn{3}{c}{$100 \cdot$ \success$\uparrow$} & \multicolumn{3}{c}{$100\ \cdot$ \fixedstrict$\uparrow$} & \multicolumn{3}{c}{\energy$\downarrow$} & \multicolumn{3}{c}{\textsc{\#Changed}$\updownarrow$} \\
Experiment  &                            Train &  Val. &  Test &                                 Train &  Val. &  Test &                 Train &  Val. &  Test &                              Train & Val. & Test \\
\midrule
1-Phase (Simple, IL)               &                            2.2 &   1.3 &   1.8 &                                 7.3 &   4.7 &   4.8 &                1.17 &  1.10 &  1.08 &                              1.1 &  0.7 &  0.6 \\
1-Phase (Simple, PPO)              &                            1.8 &   2.1 &   0.7 &                                 6.7 &   6.7 &   4.6 &                0.95 &  \textbf{0.96} &  0.99 &                              0.3 &  0.4 &  0.4 \\
1-Phase (RN18, IL)         &                            \textbf{8.2} &   1.7 &   2.8 &                                \textbf{17.9} &   5.0 &   6.3 &                \textbf{0.93} &  1.14 &  1.11 &                              1.3 &  0.9 &  0.9 \\
1-Phase (RN18, PPO)        &                            1.4 &   1.5 &   1.1 &                                 6.6 &   6.0 &   5.3 &                0.94 &  \textbf{0.96} &  \textbf{0.98} &                              0.3 &  0.3 &  0.3 \\
1-Phase (RN18+ANM, IL)     &                            4.8 &   \textbf{5.2} &   \textbf{3.2} &                                12.8 &  \textbf{11.1} &   \textbf{8.9} &                1.05 &  1.05 &  1.04 &                              1.3 &  1.0 &  1.0 \\
\midrule
2-Phase (RN18, PPO+IL)     &                            1.6 &   0.5 &   0.2 &                                 4.2 &   1.2 &   0.7 &                1.10 &  \textbf{1.15} &  1.12 &                              0.6 &  0.4 &  0.4 \\
2-Phase (RN18+ANM, PPO+IL) &                            \textbf{2.3} &   \textbf{0.6} &   \textbf{0.3} &                                 \textbf{7.3} &   \textbf{1.6} &   \textbf{1.4} &                \textbf{1.09} &  \textbf{1.15} &  \textbf{1.10} &                              0.9 &  0.5 &  0.4 \\
\midrule
\midrule
Heur. Expert               &                           85.1 &  88.0 &  83.4 &                                91.2 &  93.1 &  91.2 &                0.09 &  0.07 &  0.09 &                              2.2 &  2.2 &  2.3 \\
\bottomrule
\end{tabular}

\vspace{-0.3cm}
\caption{\textbf{Results}. For each experiment, (i) we evaluate model checkpoints, saved after approximately 0, 5, \ldots, 75 million steps, on the validation set, (ii) choose the best performing (lowest avg. $\fixedstrict$) checkpoint among these, and (iii) evaluate this best validation checkpoint on the other dataset splits. $\uparrow$ and $\downarrow$ denote if larger or smaller metric values are to be preferred, $\updownarrow$ denotes a metric that is meant to highlight behavior rather than a measure quality.}\label{tab:results}
\vspace{-0.3cm}
\end{table*}

    \noindent$\bullet$ \textsc{RGB} -- An egocentric 224${\times}$224${\times}$3 RGB image corresponding to the agent's current viewpoint (90$^\circ$ FOV). In the \Onephase task this corresponds to the RGB image from the unshuffle stage.\\
    \noindent$\bullet$ \textsc{WalkthroughRGB} -- This sensor is only available in the \Onephase task and is identical to \textsc{RGB} except it shows the egocentric image as though the agent was in the Walkthrough stage, \ie all objects were in their goal positions. It is this sensor that makes it possible, during the \Onephase task, for the agent to perform pixel-to-pixel comparisons between the environment as it should be in the walkthrough stage and as it is during the unshuffle stage.\\
    \noindent$\bullet$ \textsc{AgentPosition} -- The agent's position relative to its starting location (this is equivalent to the assumption of perfect egomotion estimation).\\
    \noindent$\bullet$ \textsc{InWalkthrough} -- Only relevant during the \Twophase task, this sensor returns ``true'' if the agent is currently in the walkthrough stage and otherwise returns ``false''.

\noindent\textbf{\Onephase Task} -- In this variant, the agent takes actions within the walkthrough and unshuffle stages simultaneously in lock step. That is, if the agent takes a \textsc{MoveAhead} action, the agent moves ahead in both stages simultaneously; as the agent begins in the same starting position in both stages, the agent's position will always be the same in both stages. As only navigational actions are allowed during walkthrough, all actions of type $\cA_{\text{Pickup}}\cup \cA_{\text{Open}}\cup \{\textsc{PlaceObject}\}$ are not executed by the agent in the walkthrough stage. During the unshuffle stage, the agent has access to the \textsc{RGB}, \textsc{WalkthroughRGB}, and \textsc{AgentPosition} sensors to complete its task.

\noindent\textbf{\Twophase Task} -- In this task, the agent must complete both the walkthrough and unshuffle stages sequentially. In this task has access to the \textsc{RGB}, \textsc{AgentPosition}, and \textsc{InWalkthrough} sensors.

\subsection{Training Pipeline}
\vspace{-0.2cm}
As our experimental results show, we found training models to complete the RoomR task using purely reward-based reinforcement learning methods to be extremely challenging. The difficulty remains even when using dense, shaped rewards. Thus, we have chosen to adopt a hybrid training strategy where we use the DD-PPO \cite{Wijmans2020DDPPOLN,schulman2017proximal} algorithm, a reward-based RL method, to train our agent when it is within the walkthrough stage, and an imitation learning (IL) approach, where we minimize a cross-entropy loss between the agent's policy and expert actions, is used when in the unshuffle stage. As it has been successfully employed in training agents in other embodied tasks (\eg \cite{GuptaTolaniEtAl2020}), for our IL training, we employ DAgger~\cite{dagger}. In DAgger, we begin training by forcing our agent to always take an expert's action with probability 1 and anneal this probability to 0 over the first 1Mn for the \Onephase task and 5Mn steps for the \Twophase task. Tacitly assumed in  the above is that we have access to an expert policy which can be efficiently evaluated at every state reached by our agent. Even with access to the full environment state, hand-designing an optimal, efficiently computable, expert is extremely difficult: simple considerations show that planning the agent's route is at least as difficult as the traveling salesman problem. Therefore, we do not attempt to design an optimal expert and, instead, a greedy heuristic expert with some backtracking and error detection capabilities. See Appendix \ref{sec:app-heuristic-expert} for more details. This expert is not perfect but, as seen in Tab.~\ref{tab:results}, can restore all but a small fraction of objects to their rightful places. For additional training details, see Appendix \ref{sec:app-implementation-details}.

\subsection{Results}
\vspace{-0.2cm}
Recall from Sec.~\ref{sec:dataset} that our dataset contains a training set of size 4000 and validation/testing sets of 1000 instances each. We report results on each of these splits but, for efficiency, include only the first 15 rearrangement instances per room in the training set (leaving 1200 instances).

\begin{figure}[t!]
     \centering
     \begin{subfigure}[b]{0.45\textwidth}
         \centering
         \includegraphics[width=\textwidth]{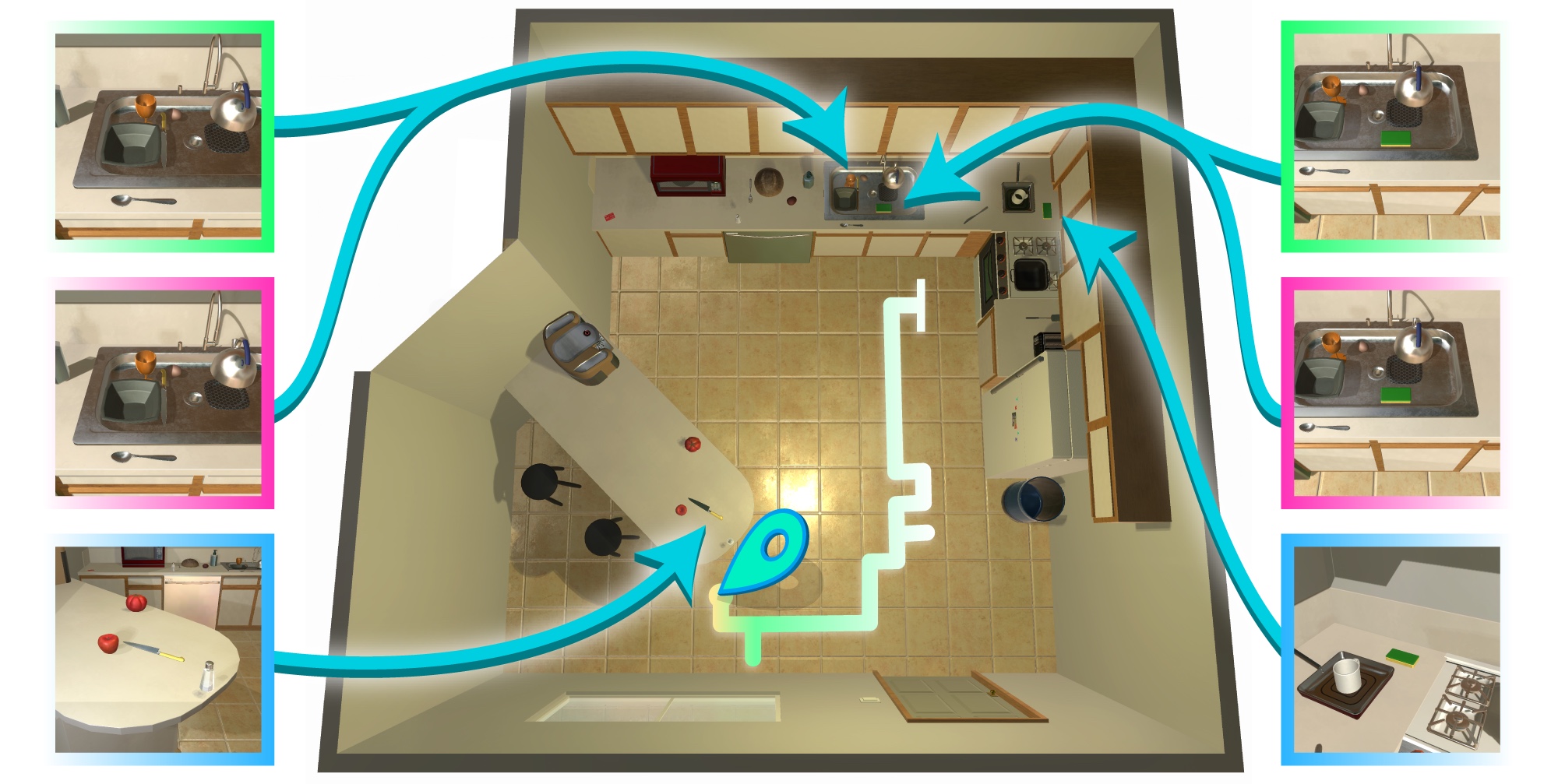}
         \caption{
         \includegraphics[height=0.09in]{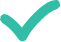}
         Successful unshuffle of a knife and dish sponge.
         }
         \label{}
     \end{subfigure}\\ 
     \begin{subfigure}[b]{0.45\textwidth}
         \vspace{-0.1cm}
         \centering
         \includegraphics[width=\textwidth]{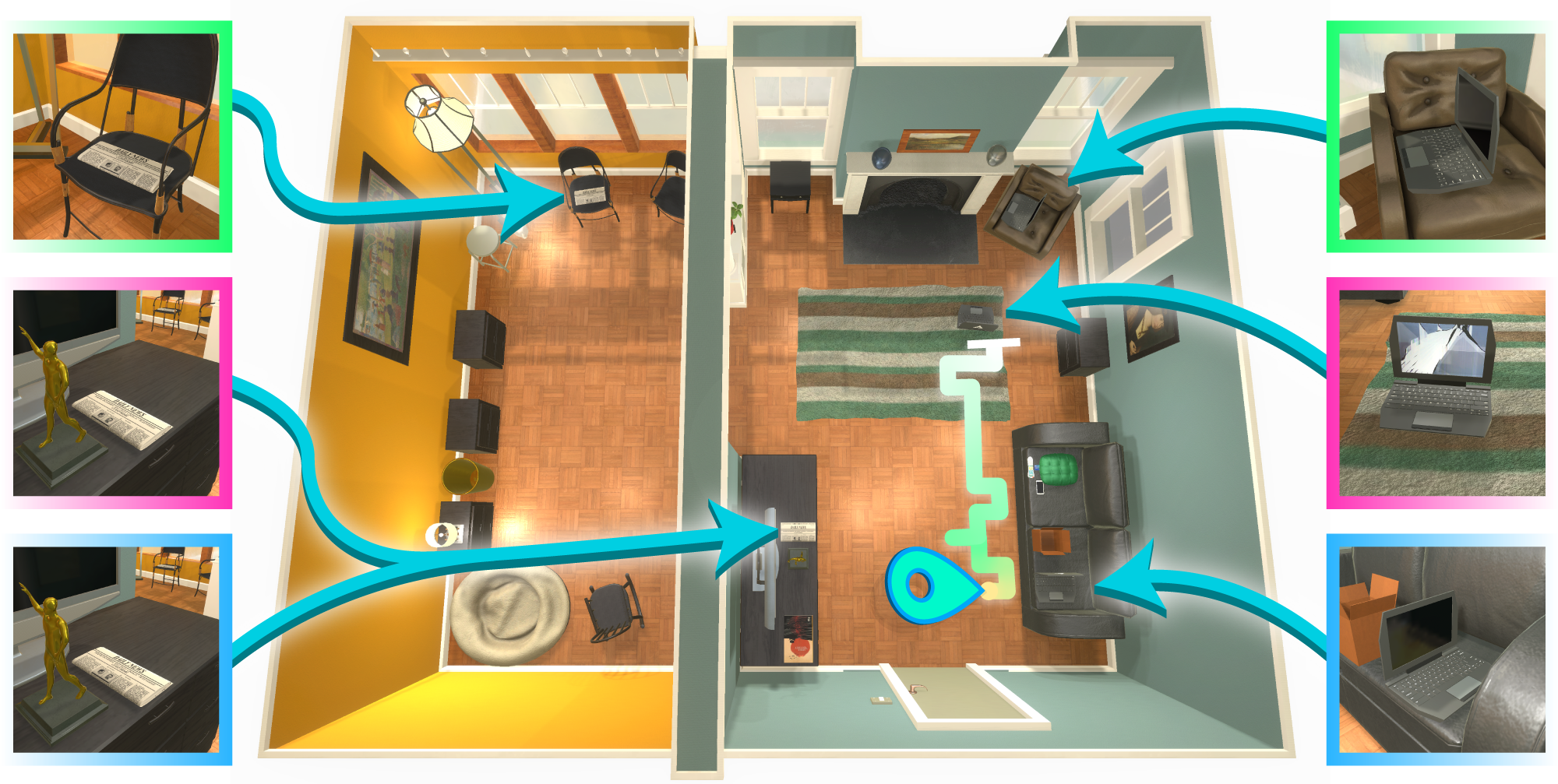}
         \caption{
         \includegraphics[height=0.09in]{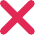}
         Unsuccessful unshuffle of a newspaper and laptop.
         }
         \label{}
     \end{subfigure}
        \vspace{-0.3cm}
        \caption{\textbf{Qualitative results.} Trajectories sampled from a \Onephase model. The goal, predicted, and initial configurations are green, pink, and blue, respectively.}
        \vspace{-0.3cm}
        \label{fig:qualitative-examples}
\end{figure}

\noindent\textbf{Baselines.} We evaluate the following baseline models: \\
\noindent \emph{$\bullet$ 1-Phase (RN18, IL)} -- An agent trained using pure imitation learning in the \Onephase task. Recall that \Onephase task models use a simplification of the model from Sec.~\ref{sec:model}, see our code for more details. \\
\noindent \emph{$\bullet$ 1-Phase (RN18, PPO)} -- As above but trained with PPO. \\
\noindent \emph{$\bullet$ 1-Phase (Simple, IL)} -- As \emph{1-Phase (RN18, IL)} but we replace the ResNet18 CNN backbone and attention module with 3 CNN blocks, this CNN is commonly used in embodied navigation baselines~\cite{Savva_2019_ICCV}. \\
\noindent \emph{$\bullet$ 1-Phase (Simple, PPO)} -- As \emph{PointU (Simple, IL)} but trained with PPO rather than IL.\\
\noindent \emph{$\bullet$ 2-Phase (RN18, PPO+IL)} -- An agent trained in the \Twophase task using the model from Sec. \ref{sec:model}. PPO and IL are used in the walkthrough and unshuffle stages, respectively.\\
\noindent \emph{$\bullet$ 1-Phase (RN18+ANM, IL)} -- We pretrain a variant of the ``Active Neural SLAM'' (ANM)~\cite{chaplot2020learning} architecture to perform semantic mapping within AI2-THOR using our set of 72 object categories. We then freeze this mapping network and train our ``1-Phase (RN18, IL)'' model extended to allow for comparing between the maps created in the unshuffle and walkthrough stages. See Appendix \ref{sec:app-semantic-mapping} for more details.\\
\noindent \emph{$\bullet$ 2-Phase (RN18+ANM, PPO+IL)} -- Similarly as above but with  semantic mapping model integrated into ``2-Phase (RN18, PPO+IL)'' baseline above.

\noindent\textbf{Analysis.} We record rolling metrics during training in Fig.~\ref{fig:training-curves}. After training, we evaluate our models on our three dataset splits and record the average metric values in Tab.~\ref{tab:results}. From the results, we see several clear trends. \\
\noindent\emph{Unshuffling objects is hard} -- Even when evaluated on the seen training rearrangements in the easier \Onephase task, the success of our best model is only 8.2\%. \\
\noindent\emph{$\bullet$ Reward-based RL struggles to train} -- Fig.~\ref{fig:training-curves} shows that PPO-based models quickly appear to become trapped in local optima. Tab.~\ref{tab:results} shows that the PPO agents move relatively few objects but, when they do move objects, they generally place them correctly even in test scenes. \\
\noindent\emph{$\bullet$ Pretrained CNN backbones can improve performance} -- We hypothesized that using a pretrained CNN backbone would substantially improve generalization performance given the relatively little object variety (compared with ImageNet) in our dataset. We see compelling evidence of this when comparing the performance of the ``1-Phase (Simple, IL)'' and ``1-Phase (RN18, IL)'' baselines (\success and \fixedstrict improvements across all splits). The results were more mixed for the PPO-trained baselines. \\
\noindent\emph{$\bullet$ The \Twophase task is much more difficult than the \Onephase task} -- Comparing the performance of the ``2-Phase (RN18, PPO+IL)'' and ``1-Phase (RN18, IL)'' baselines, it is clear that the \Twophase task is much more difficult than the \Onephase task. If the agent managed to explore exhaustively during the walkthrough stage then the two tasks would be effectively identical. This suggests that the observed gap is primarily driven by learning dynamics and the walkthrough agent's failure to explore exhaustively. Note that, as we select the best val{.} set model, Tab.~\ref{tab:results} may give the impression that the \Twophase baseline failed to train at all: this is not the case as we can see, in Fig.~\ref{fig:training-curves}, that the ``2-Phase (RN18, PPO+IL)'' baseline trains to almost the same training-set performance as the 1-Phase IL baselines. \\
\noindent\emph{$\bullet$ Semantic mapping appears to substantially improve performance} -- Our preliminary results suggest that semantic mapping can have a substantial impact on improving the generalization performance of rearrangement models, note that the ``+ANM'' baselines outperform their counterparts in almost all metrics, especially so on the validation and test sets. These results are preliminary as we have not carefully balanced parameter counts to ensure fair comparisons.

See Fig.~\ref{fig:qualitative-examples} for success and failure examples.

\vspace{-0.2cm}
\section{Discussion}
\vspace{-0.2cm}
Our proposed Room Rearrangement task poses a rich set of challenges, including navigation, planning, and reasoning about object poses and states. To facilitate learning for rearrangement, we propose the RoomR dataset that provides a challenging testbed in visually rich interactive environments. We show that modern deep RL methodologies obtain (test-set) performance only marginally above chance. Given the low performance of existing methods we suspect that future high-performance models will require novel architectures enabling comparative mapping (to record object positions during the walkthrough stage and compare these positions against those observed in the unshuffle stage), visual reasoning about object positions, and physics to be able to manipulate objects to their goal locations. Moreover, we require new reinforcement learning methodologies to allow the walkthrough and unshuffle stages to be trained jointly with minimum mutual interference. Given these challenges, we hope the proposed task opens up new avenues of research in the domain of Embodied AI.

{\small
\bibliographystyle{ieee_fullname}
\bibliography{egbib}
}

\clearpage

\newpage
\appendix
\section*{Appendix}



\section{Implementation details} \label{sec:app-implementation-details}

We train our agents using the AllenAct Embodied AI framework \cite{allenact} for ${\sim}$75 Mn steps. We run our experiments on \texttt{g4dn.12xlarge} Amazon EC2 instances which has 4 NVIDIA T4 GPUs and 48 CPU cores. See Table \ref{tab:hyperparams} for an accounting of our training hyperparameters (e.g. learning rate, loss weights, etc.). During training we obtain an FPS of ${\sim}$125 when training models with expert supervision and an FPS of ${\sim}300$ when training purely with PPO. Thus, training for ${\sim}$75 Mn steps requires approximately 4.3 days when using expert supervision and 1.8 without.

The reward structures for our agents differ in the walkthrough and unshuffle stages. Rather than provide explicit details here, as these are better read directly from code, we give some intuition about these reward structures. 

\noindent\textbf{Unshuffle stage rewards.} For the unshuffle stage the reward is quite simple. Suppose that before the agent takes an action the scene is in state $s^1\in \cS$ and, after the agent takes a step, the scene is in state $s^2\in\cS$. The agent's reward is then equal to the change in energy of the scene (with respect to the goal pose $s^*$), i.e. $D(s^1,s^*)-D(s^2,s^*)$. Thus if the energy has decreased ($D(s^2,s^*)<D(s^1,s^*)$) so that the scene is closer to the goal state than it was before, then the agent gets a positive reward. Otherwise, the agent may receive a negative reward. At the end of an unshuffle episode the agent receives a penalty equal to the negation of the remaining energy.

\noindent\textbf{Walkthrough stage rewards.} In the walkthrough stage we would like the agent to see as many of the objects in the scene as possible so that, during the unshuffle stage, the agent can compare the object poses seen against their goal positions. To this end, after every step in the walkthrough stage, the agent receives reward if it observes objects that it has never seen previously in the episode. At the end of the episode we provide the agent a reward based on the proportion of objects the agent has seen among all objects in the scene. We found this reward helpful to encourage the agent to be as exhaustive as possible.

\begin{table}[]
    \centering
    \setlength{\tabcolsep}{10pt}
    \begin{tabular}{ll}
        \hline\hline
        \textbf{Hyperparamter} &  \textbf{Value}\\\hline
        \hline\multicolumn{2}{c}{\textit{PPO}}\\\hline
        Discount factor ($\gamma$) & $0.99$\\
        GAE parameter ($\lambda$) & $0.95$\\
        Value loss coefficient & $0.5$\\
        Entropy loss coefficient & $0.01$\\
        Clip parameter ($\epsilon$)~\cite{schulman2017proximal} & $0.1$\\
        Decay on $\epsilon$ & $\mathtt{Linear}(1,0.39$, 75e6)\\
        \hline\multicolumn{2}{c}{\textit{PPO}-only -- Training}\\\hline
        \# Processes to sample steps & $40$ (5 per GPU)\\
        LR Decay & $\mathtt{Linear}(1,1/3,$ 25e6)\\
        \hline\multicolumn{2}{c}{\textit{IL} and \textit{IL+PPO} -- Training}\\\hline
        \# Processes to sample steps & $40$ (5 per GPU)\\
        \hline\multicolumn{2}{c}{Common -- \textit{Training}}\\\hline
        Rollout timesteps & $64$\\
        Rollouts per minibatch & $40$ \\
        Epochs & $3$\\
        Learning rate & 3e-4\\ 
        Optimizer & Adam~\cite{KingmaICLR2015adam}\\
        $(\beta_{1}, \beta_{2})$ for Adam & $(0.9, 0.999)$\\
        Gradient clip norm & $0.5$ \\
        Training steps & $75$ Million\\
        \hline\hline
    \end{tabular}
    \caption{\textbf{Training hyperparameters.} Here $\mathtt{Linear}(a, b, c)$ corresponds to linear interpolation between $a$ and $b$ within $c$ training steps.} \label{tab:hyperparams}
\end{table}

\section{Heuristic Expert}\label{sec:app-heuristic-expert}

The sole purpose of our expert is to produce expert actions for our learning agents to imitate. As such it is allowed to ``cheat'' by using extensive ground truth state information including the scene layout and poses of all objects in current and goal states. As it does not have to reason from visual input, the heuristic expert's performance cannot be fairly compared against the other agents. At a high-level our expert operates by looping through (1) selecting the closest object that is not in its goal pose, (2) navigating to this object via shortest paths computed on the scene layout, (3) picking up the object, (4) navigating to the closest position from which the object can be placed in its goal pose, and (5) placing the object. As AI2-THOR is physics based, it is possible for the above steps to fail (e.g. an object falls in the way of the agent as it navigates), because of this the agent has backtracking capabilities to allow it to give up on placing an object temporarily in the hope that, in placing other objects, it will remove the obstruction. 

\section{Lower-level actions}
\label{app:lowlevel}
As discussed in Sec. \ref{sec:roomr-actions}, in our experiments we use a ``high-level'' action space in line with prior work. We suspect (and hope) that within the next few years the rearrangement task will be solved using these high-level actions enabling us to move to low-level actions which are more easily implementable on existing robotic hardware. In preparation for this eventuality, we have implemented a number of lower-level actions. Rather than describe these actions individually, we will describe them in contrast to their higher-level counterparts.

\noindent\textbf{Continuous navigation.} In our experiments the agent moves at increments of 0.25 meters, uses 90$^\circ$ rotations, and changes its camera angle by $30^\circ$ at a time. We have implemented fully continuous motion so that the agent can rotate and move arbitrary degrees and distances respectively.\\

\noindent\textbf{Object manipulation.} Our high-level actions include a \textsc{PlaceObject} action that abstracts away the subtleties of moving a held object to a goal location. In our low-level actions we now allow the agent to move a held object through space (within some distance of the agent) possibly colliding with other objects. The agent then must explicitly drop the object into to the goal location.\\

\begin{figure}[b!]
    \centering
    \includegraphics[width=0.475\textwidth]{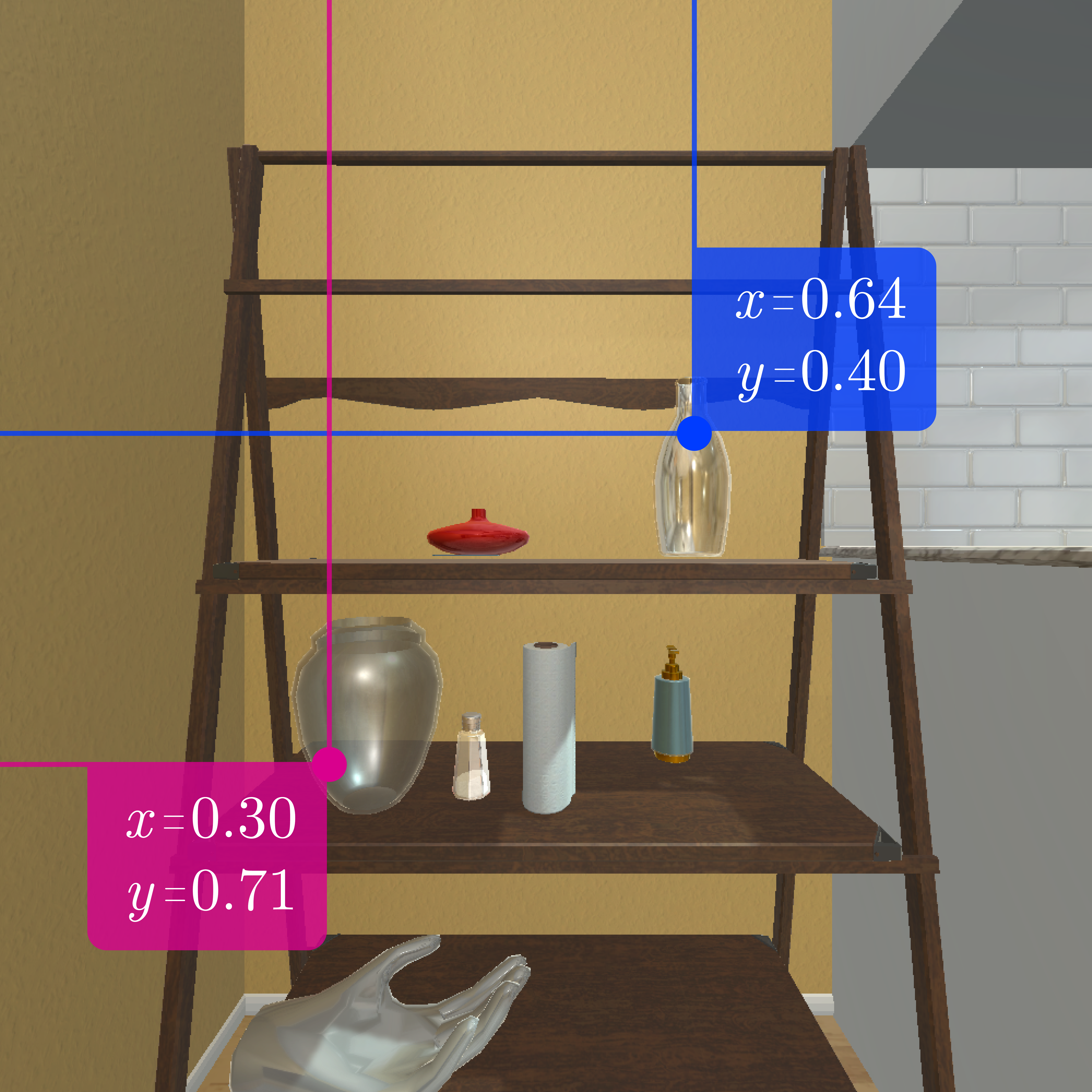}
    \caption{\textbf{Lower-level object targeting.} Instead of targeting objects based on their annotated type, the lower-level targeting action targets objects based on their location in the agent's current frame. For each $(x, y)$ coordinate, with $0\leq x, y\leq 1$, the $x$ and $y$ coordinates denote the relative distance from the left and top of the frame, respectively.}
        \vspace{-0.4cm}
    \label{fig:coordinates}
\end{figure}

\noindent\textbf{Opening and picking up objects.} When an agent opens an object using one of the 10 high-level open actions the agent is not required to specify the target openness nor specify where, in space, the object to open resides. Fig.~\ref{fig:coordinates} shows how objects are targeted with our lower-level actions. For our low-level open action the agent must specify the $(x,y)$ coordinates (in pixel-space) of the object, as well as the amount that the object is opened. Similarly, our low level \textsc{Pickup} action requires specifying the object with $(x,y)$ coordinates rather than by the object's type.

\section{Semantic Mapping}\label{sec:app-semantic-mapping}

As discussed in the main paper, we include two baselines that incorporate the ``Active Neural SLAM'' module of Chaplot et al. (2020)~\cite{chaplot2020learning} which we have adapted (by increasing the number of output channels in the map) to perform semantic mapping. 

We pretrain the ANM module so that, given a $224{\times}224{\times}3$ image from AI2-THOR, it returns a $40{\times}40{\times}75$ tensor $M$ corresponding to an estimate of the semantic map in a 2m${\times}$2m region directly in front of the agent (3 channels are used to predict free space, the other 72 are used to predict the probability that one of our 72 rearrangement objects occupies a given map location).

After pretraining this module we freeze its weights and incorporate it into our baseline model, recall Sec. \ref{sec:model}. In particular, we remove the nonparametric map from our baseline and replace it with the ANM. During the walkthrough stage the agent constructs the semantic map and saves it. During the unshuffle stage, the agent indexes into the walkthrough map to retrieve the estimate of the egocentric semantic map for the agent's current position. It compares this walkthrough map estimate against its current map estimate through the use of an attention mechanism: the two estimates are concatenated, embedded via a CNN, and then attention is computed spatially to downsample the embeddings to a single 512-dimensional vector. This embedding is then concatenated to the input to the 1-layer LSTM (recall Sec. \ref{sec:model}) along with the usual visual and discrete embeddings.

\section{Computing the energy between two poses}
\label{app:energy}
In our discussion of the ``\% Energy Remaining'' metric (recall Sec. \ref{sec:metrics}) we deferred the definition of the energy function $D:S\times S\to[0,1]$, we define this energy function now. Let $s^i=(p^1,o^1,c^1,b^1),s^2=(p^2,o^2,c^2,b^2)\in S$ be two possible poses for an object. Then,
\begin{itemize}
    \item If $b^1=1$ or $b^2=1$ we let $D(s^1,s^2)=1$.
    \item Otherwise, if the object is openable but not pickupable, we let $D(s^1,s^2)=0$ if $|o^1-o^2|\leq 0.2$ and otherwise $D(s^1,s^2)=1$, otherwise
    \item Otherwise, if the object is pickupable, we have two cases. Suppose that $\iou(s^1,s^2) > 0$. Then we let $D(s^1,s^2)=0.5\cdot \max(0, 0.5-\iou(s^1,s^2))$. Otherwise, we let $D(s^1,s^2) = 0.5 + 0.5\cdot \min(d / 2, 1)$ where $d$ be the minimum distance between a point in $c^1$ and a point in $c^2$.
\end{itemize}
Note that $D$ decreases monotonically as poses $p^1,p^2$ come closer together.

\section{Object types}
\label{app:objects}
The list of all objects have been provided in Tab.~\ref{table:object-types}.

\begin{table*}[t]
    \centering
    \footnotesize
    \begin{tabular}{lcc}
\toprule
              Object Type [A-L] &  Openable &  Pickupable \\
\midrule
        AlarmClock &     {\color{red}{\xmark}} &        {\color{green}{\cmark}} \\
      AluminumFoil &     {\color{red}{\xmark}} &        {\color{green}{\cmark}} \\
             Apple &     {\color{red}{\xmark}} &        {\color{green}{\cmark}} \\
          ArmChair &     {\color{red}{\xmark}} &       {\color{red}{\xmark}} \\
       BaseballBat &     {\color{red}{\xmark}} &        {\color{green}{\cmark}} \\
        BasketBall &     {\color{red}{\xmark}} &        {\color{green}{\cmark}} \\
           Bathtub &     {\color{red}{\xmark}} &       {\color{red}{\xmark}} \\
      BathtubBasin &     {\color{red}{\xmark}} &       {\color{red}{\xmark}} \\
               Bed &     {\color{red}{\xmark}} &       {\color{red}{\xmark}} \\
            Blinds &      {\color{green}{\cmark}} &       {\color{red}{\xmark}} \\
              Book &      {\color{green}{\cmark}} &        {\color{green}{\cmark}} \\
             Boots &     {\color{red}{\xmark}} &        {\color{green}{\cmark}} \\
            Bottle &     {\color{red}{\xmark}} &        {\color{green}{\cmark}} \\
              Bowl &     {\color{red}{\xmark}} &        {\color{green}{\cmark}} \\
               Box &      {\color{green}{\cmark}} &        {\color{green}{\cmark}} \\
             Bread &     {\color{red}{\xmark}} &        {\color{green}{\cmark}} \\
       ButterKnife &     {\color{red}{\xmark}} &        {\color{green}{\cmark}} \\
                CD &     {\color{red}{\xmark}} &        {\color{green}{\cmark}} \\
           Cabinet &      {\color{green}{\cmark}} &       {\color{red}{\xmark}} \\
            Candle &     {\color{red}{\xmark}} &        {\color{green}{\cmark}} \\
         CellPhone &     {\color{red}{\xmark}} &        {\color{green}{\cmark}} \\
             Chair &     {\color{red}{\xmark}} &       {\color{red}{\xmark}} \\
             Cloth &     {\color{red}{\xmark}} &        {\color{green}{\cmark}} \\
     CoffeeMachine &     {\color{red}{\xmark}} &       {\color{red}{\xmark}} \\
       CoffeeTable &     {\color{red}{\xmark}} &       {\color{red}{\xmark}} \\
        CounterTop &     {\color{red}{\xmark}} &       {\color{red}{\xmark}} \\
        CreditCard &     {\color{red}{\xmark}} &        {\color{green}{\cmark}} \\
               Cup &     {\color{red}{\xmark}} &        {\color{green}{\cmark}} \\
          Curtains &     {\color{red}{\xmark}} &       {\color{red}{\xmark}} \\
              Desk &     {\color{red}{\xmark}} &       {\color{red}{\xmark}} \\
          DeskLamp &     {\color{red}{\xmark}} &       {\color{red}{\xmark}} \\
           Desktop &     {\color{red}{\xmark}} &       {\color{red}{\xmark}} \\
       DiningTable &     {\color{red}{\xmark}} &       {\color{red}{\xmark}} \\
        DishSponge &     {\color{red}{\xmark}} &        {\color{green}{\cmark}} \\
            DogBed &     {\color{red}{\xmark}} &       {\color{red}{\xmark}} \\
            Drawer &      {\color{green}{\cmark}} &       {\color{red}{\xmark}} \\
           Dresser &     {\color{red}{\xmark}} &       {\color{red}{\xmark}} \\
          Dumbbell &     {\color{red}{\xmark}} &        {\color{green}{\cmark}} \\
               Egg &     {\color{red}{\xmark}} &        {\color{green}{\cmark}} \\
            Faucet &     {\color{red}{\xmark}} &       {\color{red}{\xmark}} \\
             Floor &     {\color{red}{\xmark}} &       {\color{red}{\xmark}} \\
         FloorLamp &     {\color{red}{\xmark}} &       {\color{red}{\xmark}} \\
         Footstool &     {\color{red}{\xmark}} &        {\color{green}{\cmark}} \\
              Fork &     {\color{red}{\xmark}} &        {\color{green}{\cmark}} \\
            Fridge &      {\color{green}{\cmark}} &       {\color{red}{\xmark}} \\
        GarbageBag &     {\color{red}{\xmark}} &       {\color{red}{\xmark}} \\
        GarbageCan &     {\color{red}{\xmark}} &       {\color{red}{\xmark}} \\
         HandTowel &     {\color{red}{\xmark}} &        {\color{green}{\cmark}} \\
   HandTowelHolder &     {\color{red}{\xmark}} &       {\color{red}{\xmark}} \\
        HousePlant &     {\color{red}{\xmark}} &       {\color{red}{\xmark}} \\
            Kettle &      {\color{green}{\cmark}} &        {\color{green}{\cmark}} \\
          KeyChain &     {\color{red}{\xmark}} &        {\color{green}{\cmark}} \\
             Knife &     {\color{red}{\xmark}} &        {\color{green}{\cmark}} \\
             Ladle &     {\color{red}{\xmark}} &        {\color{green}{\cmark}} \\
            Laptop &      {\color{green}{\cmark}} &        {\color{green}{\cmark}} \\
     LaundryHamper &      {\color{green}{\cmark}} &       {\color{red}{\xmark}} \\
           Lettuce &     {\color{red}{\xmark}} &        {\color{green}{\cmark}} \\
       LightSwitch &     {\color{red}{\xmark}} &       {\color{red}{\xmark}} \\
\bottomrule \\
& &
\end{tabular}
\hspace{10mm}
\begin{tabular}{lcc}
\toprule
    Object Type [M-Z] &  Openable &  Pickupable \\
\midrule
         Microwave &      {\color{green}{\cmark}} &       {\color{red}{\xmark}} \\
            Mirror &     {\color{red}{\xmark}} &       {\color{red}{\xmark}} \\
               Mug &     {\color{red}{\xmark}} &        {\color{green}{\cmark}} \\
         Newspaper &     {\color{red}{\xmark}} &        {\color{green}{\cmark}} \\
           Ottoman &     {\color{red}{\xmark}} &       {\color{red}{\xmark}} \\
          Painting &     {\color{red}{\xmark}} &       {\color{red}{\xmark}} \\
               Pan &     {\color{red}{\xmark}} &        {\color{green}{\cmark}} \\
    PaperTowelRoll &     {\color{red}{\xmark}} &        {\color{green}{\cmark}} \\
               Pen &     {\color{red}{\xmark}} &        {\color{green}{\cmark}} \\
            Pencil &     {\color{red}{\xmark}} &        {\color{green}{\cmark}} \\
      PepperShaker &     {\color{red}{\xmark}} &        {\color{green}{\cmark}} \\
            Pillow &     {\color{red}{\xmark}} &        {\color{green}{\cmark}} \\
             Plate &     {\color{red}{\xmark}} &        {\color{green}{\cmark}} \\
           Plunger &     {\color{red}{\xmark}} &        {\color{green}{\cmark}} \\
            Poster &     {\color{red}{\xmark}} &       {\color{red}{\xmark}} \\
               Pot &     {\color{red}{\xmark}} &        {\color{green}{\cmark}} \\
            Potato &     {\color{red}{\xmark}} &        {\color{green}{\cmark}} \\
     RemoteControl &     {\color{red}{\xmark}} &        {\color{green}{\cmark}} \\
         RoomDecor &     {\color{red}{\xmark}} &       {\color{red}{\xmark}} \\
              Safe &      {\color{green}{\cmark}} &       {\color{red}{\xmark}} \\
        SaltShaker &     {\color{red}{\xmark}} &        {\color{green}{\cmark}} \\
        ScrubBrush &     {\color{red}{\xmark}} &        {\color{green}{\cmark}} \\
             Shelf &     {\color{red}{\xmark}} &       {\color{red}{\xmark}} \\
      ShelvingUnit &     {\color{red}{\xmark}} &       {\color{red}{\xmark}} \\
     ShowerCurtain &      {\color{green}{\cmark}} &       {\color{red}{\xmark}} \\
        ShowerDoor &      {\color{green}{\cmark}} &       {\color{red}{\xmark}} \\
       ShowerGlass &     {\color{red}{\xmark}} &       {\color{red}{\xmark}} \\
        ShowerHead &     {\color{red}{\xmark}} &       {\color{red}{\xmark}} \\
         SideTable &     {\color{red}{\xmark}} &       {\color{red}{\xmark}} \\
              Sink &     {\color{red}{\xmark}} &       {\color{red}{\xmark}} \\
         SinkBasin &     {\color{red}{\xmark}} &       {\color{red}{\xmark}} \\
           SoapBar &     {\color{red}{\xmark}} &        {\color{green}{\cmark}} \\
        SoapBottle &     {\color{red}{\xmark}} &        {\color{green}{\cmark}} \\
              Sofa &     {\color{red}{\xmark}} &       {\color{red}{\xmark}} \\
           Spatula &     {\color{red}{\xmark}} &        {\color{green}{\cmark}} \\
             Spoon &     {\color{red}{\xmark}} &        {\color{green}{\cmark}} \\
       SprayBottle &     {\color{red}{\xmark}} &        {\color{green}{\cmark}} \\
            Statue &     {\color{red}{\xmark}} &        {\color{green}{\cmark}} \\
             Stool &     {\color{red}{\xmark}} &       {\color{red}{\xmark}} \\
       StoveBurner &     {\color{red}{\xmark}} &       {\color{red}{\xmark}} \\
         StoveKnob &     {\color{red}{\xmark}} &       {\color{red}{\xmark}} \\
           TVStand &     {\color{red}{\xmark}} &       {\color{red}{\xmark}} \\
     TableTopDecor &     {\color{red}{\xmark}} &        {\color{green}{\cmark}} \\
         TeddyBear &     {\color{red}{\xmark}} &        {\color{green}{\cmark}} \\
        Television &     {\color{red}{\xmark}} &       {\color{red}{\xmark}} \\
      TennisRacket &     {\color{red}{\xmark}} &        {\color{green}{\cmark}} \\
         TissueBox &     {\color{red}{\xmark}} &        {\color{green}{\cmark}} \\
           Toaster &     {\color{red}{\xmark}} &       {\color{red}{\xmark}} \\
            Toilet &      {\color{green}{\cmark}} &       {\color{red}{\xmark}} \\
       ToiletPaper &     {\color{red}{\xmark}} &        {\color{green}{\cmark}} \\
 ToiletPaperHanger &     {\color{red}{\xmark}} &       {\color{red}{\xmark}} \\
            Tomato &     {\color{red}{\xmark}} &        {\color{green}{\cmark}} \\
             Towel &     {\color{red}{\xmark}} &        {\color{green}{\cmark}} \\
       TowelHolder &     {\color{red}{\xmark}} &       {\color{red}{\xmark}} \\
     VacuumCleaner &     {\color{red}{\xmark}} &       {\color{red}{\xmark}} \\
              Vase &     {\color{red}{\xmark}} &        {\color{green}{\cmark}} \\
             Watch &     {\color{red}{\xmark}} &        {\color{green}{\cmark}} \\
       WateringCan &     {\color{red}{\xmark}} &        {\color{green}{\cmark}} \\
            Window &     {\color{red}{\xmark}} &       {\color{red}{\xmark}} \\
        WineBottle &     {\color{red}{\xmark}} &        {\color{green}{\cmark}} \\
\bottomrule
\end{tabular}
    \caption{\textbf{Object types.} All object types available in AI2-THOR (and thus present in our task) along with whether they are openable or pickupable.}
    \label{table:object-types}
\end{table*}

\end{document}